\documentclass[conference]{IEEEtran}
\IEEEoverridecommandlockouts
\usepackage{cite}
\usepackage{amsmath,amssymb,amsfonts}
\usepackage{algorithmic}
\usepackage{graphicx}
\usepackage{textcomp}
\usepackage{xcolor}
\usepackage{fancyhdr}
\usepackage{url}
\def\BibTeX{{\rm B\kern-.05em{\sc i\kern-.025em b}\kern-.08em
    T\kern-.1667em\lower.7ex\hbox{E}\kern-.125emX}}

\usepackage{algorithm}

\usepackage[OT2,OT1]{fontenc}
\usepackage[english]{babel}
\DeclareTextCompositeCommand{\v}{OT2}{C}{Ch}
\DeclareTextCompositeCommand{\v}{OT2}{c}{ch}
\DeclareTextCompositeCommand{\v}{OT2}{S}{Sh}
\DeclareTextCompositeCommand{\v}{OT2}{s}{sh}
\DeclareTextCompositeCommand{\v}{OT2}{Z}{Zh}
\DeclareTextCompositeCommand{\v}{OT2}{z}{zh}
\DeclareTextCompositeCommand{\'}{OT2}{C}{C1}
\DeclareTextCompositeCommand{\'}{OT2}{c}{c1}
\DeclareTextCommand{\DJ}{OT2}{Dj}
\DeclareTextCommand{\dj}{OT2}{dj}
\DeclareTextCommand{\DZ}{OT1}{Dž}
\DeclareTextCommand{\DZ}{OT2}{D2}
\DeclareTextCommand{\dz}{OT1}{dž}
\DeclareTextCommand{\dz}{OT2}{d2}

\makeatletter
\g@addto@macro\@uclclist{\dj\DJ\dz\DZ}
\makeatother

\begin{document}

\title{Intrinsically motivated option learning:\\a comparative study of recent methods
}

\author{\IEEEauthorblockN{Đorđe Božić}
\IEEEauthorblockA{\textit{School of Electrical Engineering}\\
\textit{University of Belgrade, Serbia}\\
\url{djordjebbozic@gmail.com}}
\and
\IEEEauthorblockN{Predrag Tadić}
\IEEEauthorblockA{\textit{School of Electrical Engineering}\\
\textit{University of Belgrade, Serbia}\\
\url{ptadic@etf.bg.ac.rs}}
\and
\IEEEauthorblockN{Mladen Nikolić}
\IEEEauthorblockA{\textit{Faculty of Mathematics} \\
\textit{University of Belgrade, Serbia}\\
\url{nikolic@matf.bg.ac.rs}}
}


\maketitle
\begin{abstract}

Options represent a framework for reasoning across multiple time scales in reinforcement learning (RL). With the recent active interest in the unsupervised learning paradigm in the RL research community, the option framework was adapted to utilize the concept of \textit{empowerment}, which corresponds to the amount of influence the agent has on the environment and its ability to perceive this influence, and which can be optimized without any supervision provided by the environment's reward structure. Many recent papers modify this concept in various ways achieving commendable results. Through these various modifications, however, the initial context of empowerment is often lost. In this work we offer a comparative study of such papers through the lens of the original empowerment principle.
\end{abstract}


\section{Introduction}

In the reinforcement learning (RL) paradigm, the agents select actions at each time step, which are then enacted within an environment. The agents are then given feedback in the form of the next state and the reward \cite{sutton_reinforcement_2018}. The option framework \cite{sutton_between_1999, bacon_option-critic_2017} represents an extension to the standard RL paradigm, allowing for reasoning across multiple time-scales. Here, the agents select temporally extended courses of actions, which are called options. Empowerment \cite{salge_empowerment_2013} roughly translates into the agent's ability to cover large parts of the state space and reliably infer which sequence of actions has lead them to a particular state. It was introduced to the option framework by \cite{gregor_variational_2019}, equipping the agents with intrinsic motivation and independence from the environment-based reward structure. While Gregor et al.\ \cite{gregor_variational_2019} reported excellent results on Grid World examples, they had trouble translating the concept to continuous state and action space domains when non-linear function approximators had to be utilized. Some recent papers \cite{eysenbach_diversity_2018, achiam_variational_2018} successfully applied modified versions of both the empowerment principle and the option framework to challenging continuous control domains \cite{brockman_openai_2016}, but had to significantly reframe the original problem. Certain studies \cite{aubret_survey_2019, campos_explore_2020}, however, call into question the effectiveness of the solutions proposed by \cite{eysenbach_diversity_2018, achiam_variational_2018}. 

In this paper we aim to test claims made by \cite{aubret_survey_2019, campos_explore_2020}, and offer a comparative study of \cite{gregor_variational_2019, eysenbach_diversity_2018, achiam_variational_2018} through the lens of the original empowerment principle. We first present a theoretical comparison of \cite{gregor_variational_2019, eysenbach_diversity_2018, achiam_variational_2018} in Section \ref{the}. We then outline the critique of \cite{eysenbach_diversity_2018, achiam_variational_2018} as reported in \cite{aubret_survey_2019, campos_explore_2020} in Section \ref{mot}, and detail the purpose of this paper. We present the results of our experiments with \cite{eysenbach_diversity_2018} in Section \ref{res}, and discuss them with respect to both \cite{aubret_survey_2019, campos_explore_2020}, and the original empowerment principle \cite{salge_empowerment_2013, gregor_variational_2019} in Section \ref{disc}. We conclude this paper with promising approaches for future study in Section \ref{conc}.

\section{Theoretical background}\label{the}

Empowerment \cite{salge_empowerment_2013} has been studied throughout various fields, ranging from behavioral science, across evolutionary biology to artificial intelligence. According to \cite{salge_empowerment_2013}, empowerment encapsulates both the amount of influence the agent can exert on the environment, as well as how much of that influence it can perceive. Compared to standard RL state values, which encapsulate the environment-provided reward benefit, it shares the same exploitative nature, but instead measures state value by the agent's ability to reliably influence its own future. It is thus fundamentally a state value estimate, which has the benefit of having the success criterion independent of any particular environment-defined reward structure.

Salge et al.\ \cite{salge_empowerment_2013} originally express empowerment as the mutual information (MI) between the agent's actuators and percepts obtained by acting out the actuator output, and assume an open-loop policy, whereby the agent chooses its actions with no regard to the current state. Gregor er al. \cite{gregor_variational_2019} introduce the concept of empowerment to the closed-loop policy RL setting, and replace actuators with options $\Omega$, and percepts with terminating states $s_f$. Empowerment is formalised by (\ref{eqn:mi_gregor}). It should express how much influence the agent has over the environment, which (\ref{eqn:mi_gregor}) achieves by maximizing the entropy of reachable final states $s_f$ given that the option was initiated in $s_0$, $\mathcal{H}(s_f|s_0)$. Agents should further be able to determine which option has brought them to $s_f$, corresponding to the conditional entropy $\mathcal{H}(s_f|\Omega, s_0)$ minimization. The final objective of \cite{gregor_variational_2019} is to maximize the MI
\begin{subequations}\label{eqn:mi_gregor}
\begin{align}
    \mathcal{I}(\Omega;s_f \vert s_0) 
    & = \mathcal{H}(s_f|s_0) - \mathcal{H}(s_f|\Omega, s_0) \\ 
    & =  \mathcal{H}(\Omega|s_0) - \mathcal{H}(\Omega|s_f, s_0) \label{eqn:mi_gregor_b}
\end{align}
\end{subequations}

Gregor et al.\ \cite{gregor_variational_2019} exploit the fact that mutual information is symmetric to obtain \eqref{eqn:mi_gregor_b}, which contains the conditional probability of options given states that can be interpreted as a policy. More specifically, $\mathcal{H}(\Omega|s_0)$ in \eqref{eqn:mi_gregor_b} involves the controllability distribution $p^C(\Omega|s_0)$ \cite{gregor_variational_2019}, which roughly corresponds to the policy over options \cite{sutton_between_1999} that reasons in a coarser temporal domain. Once an option $\Omega$ is drawn from $p^C(\Omega|s_0)$, the corresponding intra-option policy $\pi(a|s, \Omega)$ is followed until $s_f$ is reached. Together with the option-conditioned transition model $p^J(s_f|\Omega, s_0)$ it allows for drawing a close connection to the policy gradient theorem \cite{sutton_reinforcement_2018}, which translates \eqref{eqn:mi_gregor} to the RL setting.

The empowered agent reasoning across multiple time scales proposed in \cite{gregor_variational_2019} is given in Algorithm \ref{alg:vic}. Instead of using extrinsic, environment-provided reward structure, it maximizes $r_I$, which is obtained as a direct consequence of the connection of objective (\ref{eqn:mi_gregor}) to the policy gradient theorem \cite{sutton_reinforcement_2018}, and defined in the algorithm below. Finally, both $p^C(\Omega|s_0)$ and $p(\Omega|s_f, s_0)$ are represented as neural networks with parameters $\theta$ and $\phi$, respectively, where the latter is written as $q_\phi(\Omega|s_f, s_0)$ to signify that it is formally obtained using a variational approach \cite{barber_im_nodate} applied to \eqref{eqn:mi_gregor}.

\begin{algorithm}[H]
\caption{Variational Intrinsic Control \cite{gregor_variational_2019}}\label{alg:vic}
\begin{algorithmic}
\footnotesize
\STATE Assume an agent starts in $s_0$
\FOR{episode = $1, M$}
    \STATE Sample $\Omega \sim p^C_\theta(\Omega|s_0)$
    \STATE Follow intra-option policy $\pi(a|\Omega, s)$ till termination state $s_f$
    \STATE Regress the discriminator $q_{\phi}(\Omega|s_0, s_f)$ towards chosen $\Omega$
    \STATE Calculate intrinsic reward $r_I = \log q_\phi(\Omega|s_0,s_f) - \log p^C_\theta(\Omega|s_0)$
    \STATE Use a reinforcement learning algorithm update for $\pi(a|\Omega, s)$ to maximize $r_I$ 
    \STATE Reinforce option prior $p^C_\theta(\Omega|s_0)$ based on $r_I$
    \STATE Set $s_0 = s_f$
\ENDFOR
\end{algorithmic}
\end{algorithm}
\vspace*{-1mm}

The principal issue reported in \cite{gregor_variational_2019} is the problem of exploration. Namely, the term $\log q_\phi(\Omega|s_f, s_0)$ will penalize the agent when it encounters previously unseen states, as the discriminator has not yet learned to associate those states with options that have just discovered them, actively discouraging exploration. This is not surprising however, as according to the empowerment principle agents should be aware of their influence over the environment, which in this concrete realization of empowerment refers to being able to determine options from states they visit. According to the other postulate of empowerment, agents should maximize their influence over the environment, which should push the agents to explore in order to balance out the previous penalty. This should be achieved by maximizing the second term $-\log p_\theta^C(\Omega|s_0)$, but according to \cite{gregor_variational_2019} this does not happen when non-linear function approximators are used to estimate intrinsic reward terms. This is because both estimators would need to have converged in order to adequately represent the true empowerment objective, which is not the case during training. The intrinsic reward is thus a noisy learning signal and leads to unstable training. 

In \cite{eysenbach_diversity_2018,achiam_variational_2018} the modified empowerment principle is relatively successfully applied to challenging continuous control tasks \cite{brockman_openai_2016}, with the introduction of a fixed uniform prior $p(\Omega)$ instead of the learned $p^C_\theta(\Omega|s_0)$. This solves both problems reported in \cite{gregor_variational_2019}, as the controllability distribution is no longer a neural network and assures the maximum entropy of the second term in $r_I$. Unfortunately, this also completely changes the original motivation that has lead to the introduction of options \cite{sutton_between_1999}. Agents no longer reason on multiple time-scales, as no reasonable composition of options can be achieved since $p(\Omega)$ has no state dependence and is never reinforced. Solutions proposed by \cite{eysenbach_diversity_2018, achiam_variational_2018} diverge from this point into two separate directions, which we present in the following two subsections.

\subsection{Diversity Is All You Need (DIAYN)} 

Eysenbach et al.\ \cite{eysenbach_diversity_2018} additionally employ the maximum entropy framework \cite{pmlr-v70-haarnoja17a} to improve the exploration capability of the agent. They thus implement the intra-option policies using the Soft Actor Critic (SAC) algorithm \cite{haarnoja_soft_2018}, which leads to the modified empowerment objective \eqref{eqn:mi_diayn}. The entropy term $\mathcal{H}(a|\Omega, s)$ encourages intra-option policies to act as stochastically as possible, while the rest of the objective corresponds to (\ref{eqn:mi_gregor}) and assures that options are indeed discriminable. The final objective maximized in \cite{eysenbach_diversity_2018} is thus:
\begin{equation} \label{eqn:mi_diayn}
    \mathcal{I}(\Omega; s) + \mathcal{H}(a|\Omega, s)
\end{equation}

The second modification they introduce is that options are no longer discriminated solely based on the starting and ending states $(s_0, s_f)$, but rather with respect to every state $s$ generated by the option. Together, these two modifications enforce stronger option diversity, and yield the intrinsic reward
\begin{equation} \label{eqn:diayn_intrinsic}
    r_I = \log q_{\phi}(\Omega|s) - \log p(\Omega)
\end{equation}
which does not convey the intra-option policy stochasticity incentive as it is handled as part of the internal SAC update.

\subsection{Variational Option Discovery Algorithms (VALOR)}

Achiam et al.\ \cite{achiam_variational_2018} propose an interesting new implementation for the empowerment objective drawing close connection to the variational auto-encoder \cite{kingma_auto-encoding_2014}. They feed the option $\Omega \sim p(\Omega)$ to the encoder, where the encoder is represented by a mixture of environment dynamics and intra-option policies $\pi(a|s, \Omega)$. The produced ``code'' is the episode trajectory $\tau$, upon which the decoder is trained to match the originally fed option $\Omega$. In other words, the agent encodes the option choice by acting according to the option-conditioned policy within an environment. The empowerment objective \eqref{eqn:mi_gregor} thus translates into the ability of the decoder to match the initial option choice, indicating that intra-option policies are discernible. Similarly to DIAYN, VALOR keeps the probability distribution $p(\Omega)$ fixed and state independent, and enacts single options throughout the entirety of episodes, without interruption. Since $p(\Omega)$ is chosen to be uniform, the second term of (\ref{eqn:mi_gregor}) is maximized.

\section{Motivation}\label{mot}

In \cite{eysenbach_diversity_2018, achiam_variational_2018} the initial empowerment principle is modified in two significant ways. Firstly, the controllability distribution is no longer learned and is kept uniform to maximize the term $\mathcal{H}(\Omega|s_0)$ in objective (\ref{eqn:mi_gregor}). Secondly, options uninterruptedly execute for the entire duration of episodes, allowing them to maximize the diversity of options with respect to \emph{all} states, as indicated by the term $q_\phi(\Omega|s)$ of (\ref{eqn:diayn_intrinsic}).

The efficiency of \cite{eysenbach_diversity_2018,achiam_variational_2018} is brought into question in \cite{aubret_survey_2019, campos_explore_2020}. Aubret et al.\ \cite{aubret_survey_2019} state that it is not apparent how these algorithms can learn \emph{meaningful} policies on tasks where terminating states are distinguishable, or simply do not exist. Namely, since the entropy over the controllability distribution is already preemptively maximized from the start, these agents essentially only optimize for the diversity of states that options visit. If a certain environment does not indeed have similar terminating states which indicate agent's failure to learn meaningful behavior, then optimizing purely for state diversity may not be enough to learn such behavior. Here, we fully acknowledge that the term ``meaningful'' may be vague, and discuss it in more detail in the following sections. Campos et al.\ \cite{campos_explore_2020} outline severe exploration problems demonstrating that these agents fail to leave the initial room in a maze-like setting. More specifically, they show that agents proposed by \cite{eysenbach_diversity_2018, achiam_variational_2018} exhibit behavior similar to a random exploratory policy, but are good at partitioning states such policy visits into separate options.

The purpose of this paper is two-fold. Firstly, we aim to test the claims made by \cite{aubret_survey_2019, campos_explore_2020} and to that end we implement DIAYN \cite{eysenbach_diversity_2018}, verify our implementation against the original one, and run experiments on continuous control tasks \cite{brockman_openai_2016}. Secondly, we analyze the modifications made in \cite{eysenbach_diversity_2018, achiam_variational_2018} through the lens of the original empowerment principle, and try to determine whether effects outlined in \cite{aubret_survey_2019, campos_explore_2020} may be caused by these changes.

\section{Results}\label{res}

\begin{figure}[htbp]
\centerline{\includegraphics[width=\columnwidth]{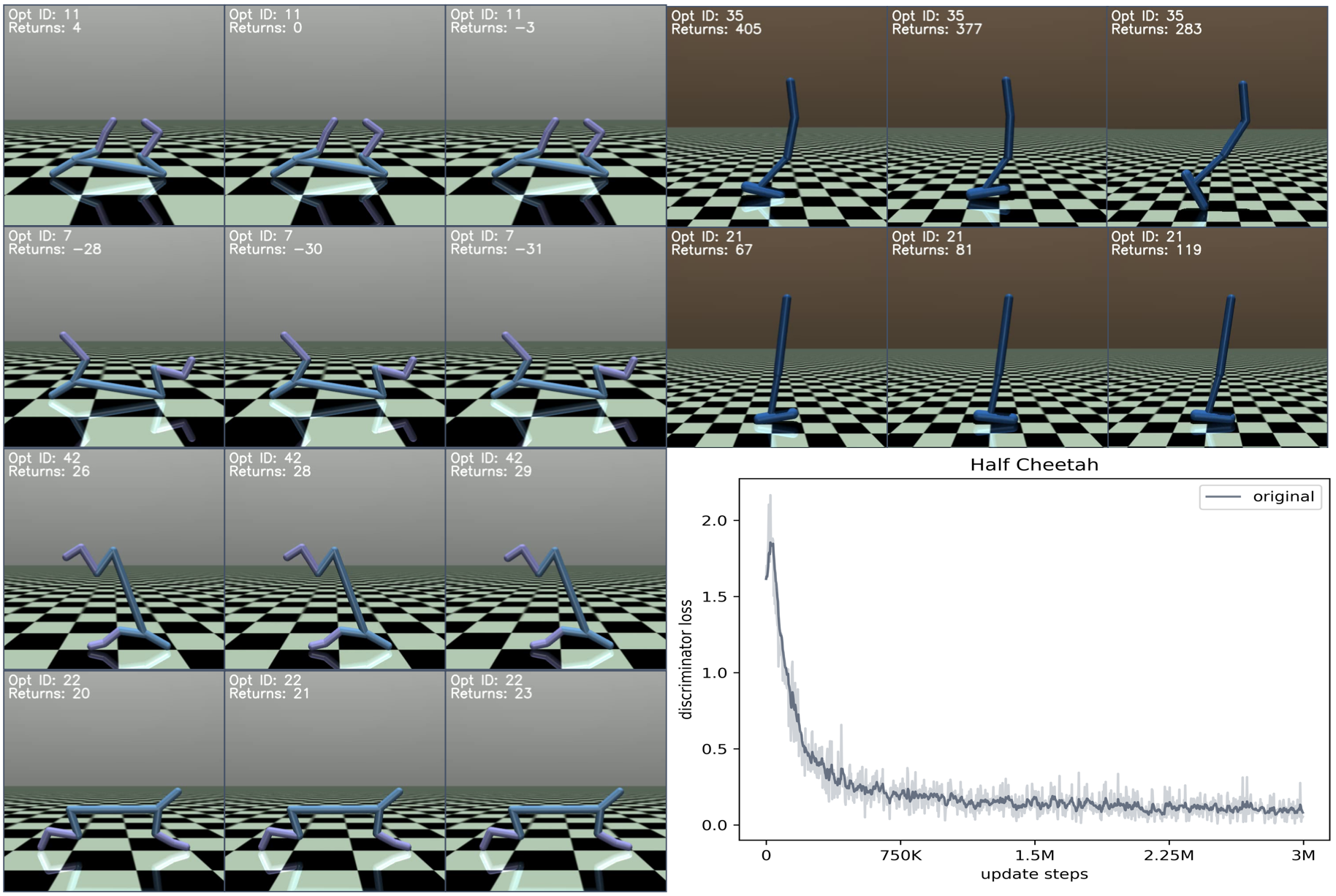}}
\caption{DIAYN option diversity in the Half Cheetah (left), and the Hopper (top right) environments. Each row contains frames obtained by randomly sampling a single episode played out with a deterministic intra-option policy. Half Cheetah discriminator $q_\phi(\Omega|s)$ loss (bottom right).}
\label{fig:diayn}
\end{figure}

We replicate the original training hyper-parameters reported in \cite{eysenbach_diversity_2018}, and successfully reproduce DIAYN's success on the Inverted Pendulum, Mountain Car, and Hopper environments. Among 50 options the agent was initialized with, most solve these environments capturing diverse behaviors. Two such behaviors are shown in the top two rows of Figure \ref{fig:diayn}, on the right. In the first row, the agent learns to take confident strides forward, resembling the actual environment-directed optimal policy. The second behavior represents tip-toeing in place, where the agent still keeps its balance successfully. Other learned options include hopping backwards, striding forwards in asymmetric steps, and many other slightly different variants.

Unfortunately, we fail to replicate DIAYN's behavior on more challenging environments such as Half Cheetah or Ant. Indeed, Eysenbach et al.\ \cite{eysenbach_diversity_2018} report that some supervision was required to make the agent learn useful behaviors on the Ant environment, but it was surprising to see that among 50 options, none learned any meaningful behaviors on the Half Cheetah task. Here we would define such behavior as that observed in Hopper or Inverted Pendulum---one that resembles the optimal policy, possibly differing slightly from it on account of diversity. Instead, in Half Cheetah, the agent learns options that opt to hold static states throughout entire episodes. Moreover, in many of these states, the agent's ability to move forward, or meaningfully change its position, is either limited, or non-existent. Figure \ref{fig:diayn} shows two such options, where the agent lays on its back. Interestingly enough, judging by the discriminator $q_\phi(\Omega|s)$ loss, shown in Figure \ref{fig:diayn}, the agent seems to have successfully solved its objective.

\section{Discussion}\label{disc}

Based on our experiments with DIAYN, concerns presented in \cite{aubret_survey_2019, campos_explore_2020} seem valid. We start this section by dissecting the observed behavior on the Half Cheetah environment through the lens of both of these works, and then present our own analysis with respect to the original empowerment principle.

According to \cite{aubret_survey_2019}, DIAYN's failure is tied to the nature of the environment itself. Namely, failure to achieve desirable behaviour does not terminate the episode in Half Cheetah, which is not the case with Inverted Pendulum and Hopper. Combined with the fact that Half Cheetah has a much richer state space compared to the other two, the agent can boost the discriminability of states visited by options without learning anything useful, as indicated by the discriminator loss in Figure \ref{fig:diayn}. In other words, assuming static but mutually distinguishable positions does indeed optimize the objective. Inverted Pendulum and Hopper environments do not permit the agent to do this, as failure to keep balance immediately terminates the episode, resulting in highly non-discriminable states across options. The discriminability objective is thus directly tied to learning meaningful behaviors in such environments.

Campos et al.\ \cite{campos_explore_2020} point to exploration as the main culprit for the poor performance of algorithms \cite{eysenbach_diversity_2018, achiam_variational_2018}. More specifically, the authors consider a uniform prior over $N$ options, as in DIAYN and VALOR, and compare intrinsic rewards $r_I = \log p(\Omega|s) - \log p(\Omega)$ for known and unseen states:
\begin{align}
    r_I^{\text{known}} &= \log 1 + \log N = \log N
    \label{eqn:r_known}
    \\
    r_I^{\text{unseen}} &= \log \frac{1}{N} + \log N = 0
    \label{eqn:r_unseen}
\end{align}

They show that DIAYN does not learn options that are able to leave the initial room in a maze-like environment, prematurely committing to already discovered states and exhibiting behavior no different than that of a random policy. This is blamed on the reward structure, which favors visiting known regions of the state space and provides no incentive for exploration. In our experiments, this behavior is not observed in Hopper and Inverted Pendulum, where agents do learn diverse behaviors, but \emph{is} observed in Half Cheetah. Namely, all of the states that options finally converge to are easily reachable right from the start by any policy. The agent progresses neither forwards nor backwards, a behavior that is not trivial to learn given the dimensionality of the action space, and most it does is to nose-dive or flip over. Both of these behaviors are achievable by accident with an untrained exploratory policy.

DIAYN's poor performance on certain environments is further exacerbated by its departure from the original empowerment principle. As we've already outlined, DIAYN and VALOR depart from \cite{salge_empowerment_2013, gregor_variational_2019} in two significant ways -- (i) the entropy of the controllability distribution is preemptively maximized, and, (ii) options are never switched during episodes and their diversity is measured with respect to each state they visit. In such a setting, the intrinsic reward term corresponding to the controllability distribution is guaranteed to consistently highly reward any behaviour the agent produces. The agent can then maximize the other part of the objective concerning option discriminability simply by meaninglessly varying elements of the state vector, since the rich state space $s \in R^{17}$ is immediately reachable. With both of the intrinsic reward terms artificially maximized, the agent is free to converge to holding either completely, or partially, unrecoverable states fixed. While the agent's ability to cleverly exploit the discriminability term is tied to the nature of the environment itself, points (i) and (ii) make the intrinsic reward less reflective of the agent’s actual performance compared to \cite{gregor_variational_2019}. If multiple options were used within episodes, all future options initialised in unrecoverable states would have been indistinguishable, garnering lower intrinsic rewards. On the other hand, for at least partially recoverable states, the learned controllability distribution would have had to resort to selecting only a handful of options to escape, obtaining low rewards due to its reduced entropy. In both cases the intrinsic rewards the agent receives would be lower, and thus more reflective of its performance compared to DIAYN. Consequentially, DIAYN’s success heavily depends on the nature of the environment---it may be unaffected as in Hopper or Inverted Pendulum, because optimizing for state discriminability is sufficient in these environments, but may break even on similar tasks, such as Half Cheetah. This observation principally agrees with \cite{aubret_survey_2019}, but relating empowerment to exploration \cite{campos_explore_2020} requires a more careful approach.

The empowered agents are interested in states for which they can identify the option that has produced them, and from which they can choose multiple options to follow. They thus have a built-in incentive to avoid getting stuck or ending episodes too soon if state discriminability is not reached, but otherwise are neither actively hindered nor enforced to explore \cite{salge_empowerment_2013}. The reward structure analysis shown in \eqref{eqn:r_known} and \eqref{eqn:r_unseen} is thus in perfect accordance with the empowerment principle---the agents should indeed visit states that are known to them, and feel neutral about states which they observe for the first time. If we were to pose the meaningless behavior DIAYN may converge to as the problem of exploration, we posit that the constant high reward the agent receives is to be blamed, as the agent has little incentive to explore further.

\section{Conclusion}\label{conc}

While the authors of \cite{eysenbach_diversity_2018, achiam_variational_2018} successfully apply empowerment inspired agents on some of the challenging continuous control tasks, our experiments show that \cite{aubret_survey_2019, campos_explore_2020} offer valid critiques. The main contribution of this work is showing that shortcomings of these algorithms, as expressed by \cite{aubret_survey_2019, campos_explore_2020}, can also be seen through the original concept of empowerment. Based on our analysis, we propose that any research done in this domain should be in accordance with the original empowerment objective \cite{salge_empowerment_2013, gregor_variational_2019}. This has the potential to avoid the exploration issue outlined in \cite{campos_explore_2020}, and the environment dependency issue presented in \cite{aubret_survey_2019}, and has the additional benefit of retaining the composite nature of options inherent to the original option framework \cite{sutton_between_1999, bacon_option-critic_2017}.

\bibliographystyle{ieeetr}
\bibliography{bozic2021.bib}

\section{Supplementary Materials}

\begin{figure}[htbp]
\centerline{\includegraphics[width=\columnwidth]{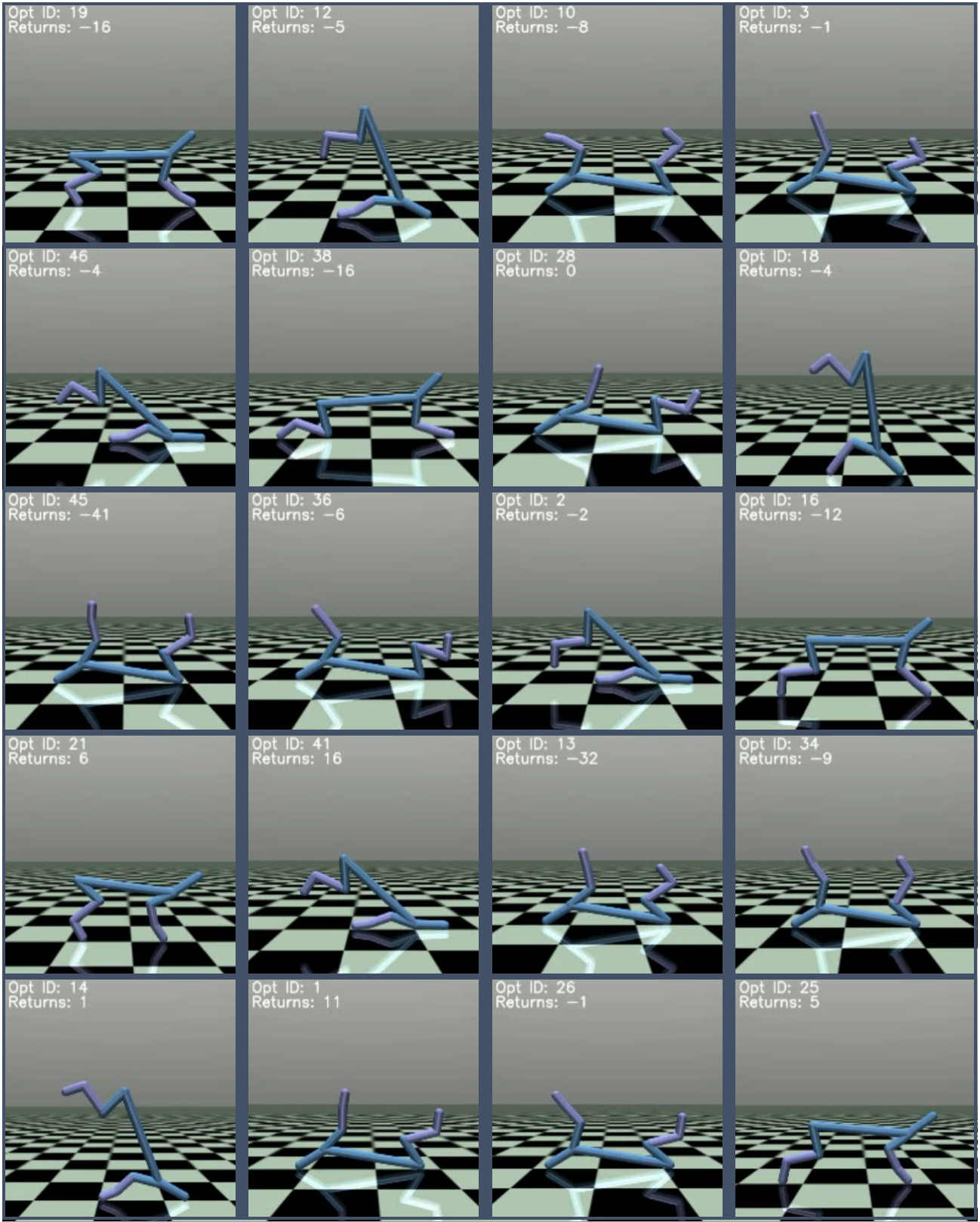}}
\caption{Random selection of 20/50 DIAYN options on Half Cheetah. Each square represents a separate option in which the agent keeps the position fixed throughout the entire episode.}
\label{fig:diayn_supp}
\end{figure}

Figure \ref{fig:diayn_supp} shows the behavior of twenty randomly sampled options, out of fifty DIAYN was initialized with. These behaviors were generated using a deterministic option policies the agent has converged towards after ten million steps. 

The presented states were reached in only a couple of steps, after which the agent chose to keep them fixed throughout the rest of the episodes. Out of these behaviors eleven are easily reached with no training whatsoever, and nine can be reliably reached after only a couple of updates to the policy. Furthermore, nine options end up in states that are completely unrecoverable, six end up in semi-recoverable states, and only five options end up in states that would be in accordance to the controllability part of the original empowerment objective.

\end{document}